\documentclass[lettersize,journal]{IEEEtran}
\usepackage{amsmath,amsfonts}
\usepackage{algorithmic}
\usepackage{algorithm}
\usepackage{array}
\usepackage[caption=false,font=normalsize,labelfont=sf,textfont=sf]{subfig}
\usepackage{textcomp}
\usepackage{stfloats}
\usepackage{url}
\usepackage{verbatim}
\usepackage{graphicx}
\usepackage{cite}
\usepackage{booktabs}

\begin{document}

\title{HGACNet: Hierarchical Graph Attention Network for Cross-Modal Point Cloud Completion}

\author{Yadan Zeng$^{1}$, Jiadong Zhou$^{1}$, Xiaohan Li$^{3}$, I-Ming Chen$^{1}$

\thanks{Y. Zeng, J. Zhou and I-M. Chen are with the Robotics Research Centre of the School of Mechanical and Aerospace Engineering, Nanyang Technological University, Singapore.
}
\thanks{X. Li is with the College of Information and Control Engineering, Xi'an University of Architecture and Technology, Xi’an, China.
}
}


\markboth{Journal of \LaTeX\ Class Files,~Vol.~14, No.~8, August~2021}%
{Shell \MakeLowercase{\textit{et al.}}: A Sample Article Using IEEEtran.cls for IEEE Journals}


\maketitle

\begin{abstract}
Point cloud completion is essential for robotic perception, object reconstruction and supporting downstream tasks like grasp planning, obstacle avoidance, and manipulation. However, incomplete geometry caused by self-occlusion and sensor limitations can significantly degrade downstream reasoning and interaction. 
To address these challenges, we propose HGACNet, a novel framework that reconstructs complete point clouds of individual objects by hierarchically encoding 3D geometric features and fusing them with image-guided priors from a single-view RGB image. At the core of our approach, the Hierarchical Graph Attention (HGA) encoder adaptively selects critical local points through graph attention-based downsampling and progressively refines hierarchical geometric features to better capture structural continuity and spatial relationships.
To strengthen cross-modal interaction, we further design a Multi-Scale Cross-Modal Fusion (MSCF) module that performs attention-based feature alignment between hierarchical geometric features and structured visual representations, enabling fine-grained semantic guidance for completion. In addition, we proposed the contrastive loss (C-Loss) to explicitly align the feature distributions across modalities, improving completion fidelity under modality discrepancy.
Finally, extensive experiments conducted on both the ShapeNet-ViPC benchmark and the YCB-Complete dataset confirm the effectiveness of HGACNet, demonstrating state-of-the-art performance as well as strong applicability in real-world robotic manipulation tasks.

\end{abstract}


\section{Introduction}
\IEEEPARstart{P}{oint} clouds, as a fundamental representation of 3D geometry, are widely used in robotics, autonomous driving, and scene reconstruction due to their ability to capture precise spatial and structural details \cite{completeSUVey1, completeSUVey2, completeSUVey3}. However, real-world point clouds obtained from LiDAR, depth cameras, and other 3D sensors often suffer from self-occlusion, sensor noise, limited resolution, and restricted field of view, leading to fragmented, sparse, or incomplete representations. Such incomplete geometry can significantly impair object recognition, pose estimation, and obstacle avoidance, ultimately degrading the reliability of downstream tasks such as robotic grasping and motion planning \cite{applicationsimulating, applicationpaint,applicationkitting, Scpnet}, where accurate 3D structure is critical for safe and effective interaction.

To address this challenge, object-level point cloud completion aims to reconstruct complete and structurally consistent 3D shapes of individual objects by inferring missing geometry. Unlike denoising or upsampling, which primarily refine existing points \cite{ pc_upsampling2}, completion explicitly recovers unobserved structures by leveraging data-driven geometric priors. While traditional single-modal methods \cite{GRnet, FoldingNet, Pointr} adopt encoder-decoder architectures to predict shapes from partial inputs, they often struggle with occlusions and lack the ability to infer fine-grained structures, especially in symmetric or complex objects. Recent cross-modal approaches incorporate RGB images to provide semantic guidance \cite{CSDN, XMFNet}. However, they often fail to fully exploit cross-modal relationships, relying on early-stage fusion or computationally intensive attention modules, which limits semantic alignment and fine-grained reconstruction quality.

\begin{figure}[!t]
\centering
\vspace{0.04cm}
\centering
\includegraphics[width=0.48\textwidth]{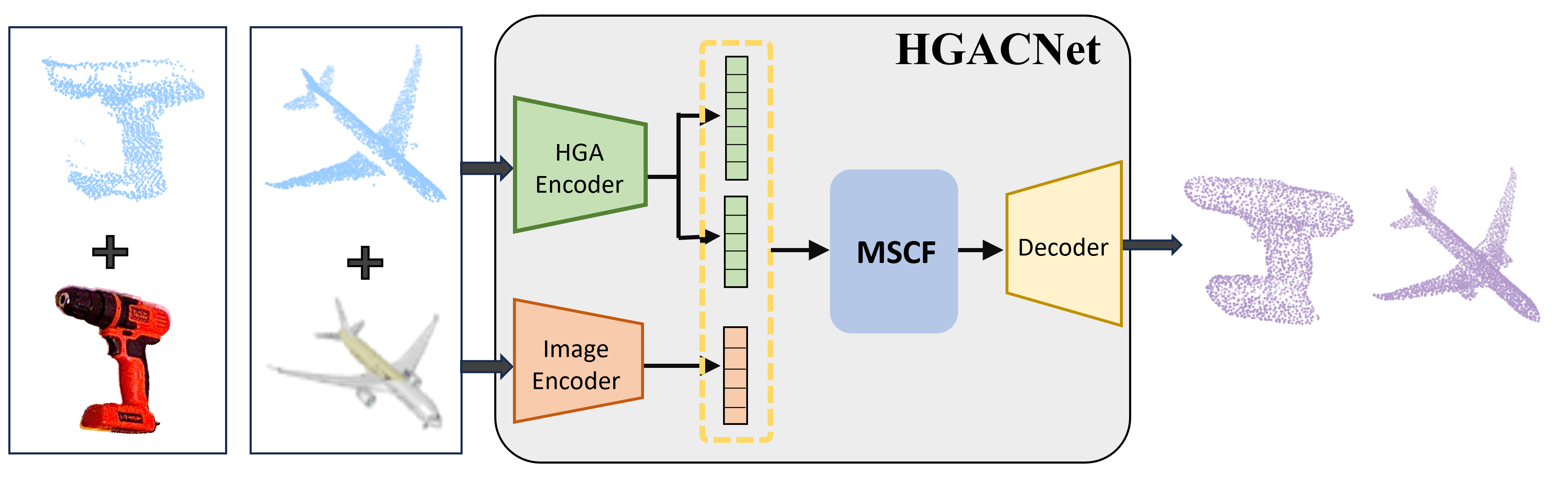}
\vspace{-0.1cm}
\caption{Overview of HGACNet. Given a partial point cloud and an RGB image, HGACNet fuses geometric and visual features to reconstruct complete 3D object shapes, and works effectively on both synthetic and real-world data.}
\label{fig1}
\vspace{-0.4cm}
\end{figure}

To overcome these limitations, we propose HGACNet, a cross-modal point cloud completion framework that hierarchically encodes 3D geometric features and fuses them with image-guided priors, as illustrated in Fig.~\ref{fig1}. Our architecture consists of two core components: the Hierarchical Graph Attention (HGA) encoder and the Multi-Scale Cross-Modal Fusion (MSCF) module.
The HGA encoder applies graph attention-based downsampling to adaptively select structurally important local points, while progressively refining hierarchical geometric features to capture both fine-grained structures and global context.
Parallel to the geometric branch, the image branch incorporates Swin Transformer \cite{SwinTransformer} for visual feature extraction. The architecture excels at capturing both fine-grained textures and scene-level context via its shifted window-based self-attention, while its multi-stage design naturally complements the hierarchical structure of our point cloud encoder.

Subsequently, the MSCF module performs both intra-modal fusion to enhance geometric consistency within the point cloud domain, and inter-modal fusion to align geometric features with structured visual representations via cross-attention, enabling instance-level alignment and fine-grained semantic guidance for completion.
To further reduce the modality gap, we propose C-Loss, a novel contrastive loss function  that enhances cross-modal feature alignment by enforcing consistency between point cloud and image modalities. Specifically, C-Loss encourages semantically corresponding features to be closer in the latent space while separating unrelated features, facilitating more coherent and accurate reconstruction.

Finally, we conduct comprehensive experiments on ShapeNet-ViPC benchmark, where HGACNet outperforms the state-of-the-art (SOTA) method EGIINet \cite{EGIINet} by 17\% in L2-CD. Additionally, evaluation on the real-world YCB-Complete dataset highlights its practical applicability in robotic perception scenarios. 

The key contributions of this work are summarized as follows:

\begin{itemize}
\item{We propose HGACNet, a hierarchical cross-modal framework that integrates hierarchical graph attention encoding with multi-scale cross-modal fusion, effectively leveraging both point cloud geometry and image priors for high-fidelity completion.}
\item{We design a HGA encoder that adaptively selects critical local points and refines geometric features to preserve structural continuity and fine-grained details.}
\item{We develop a MSCF module that enables instance-level alignment between geometric and visual representations, complemented by our proposed C-Loss to reduce modality discrepancies and improves reconstruction accuracy.}
\end{itemize}

\section{Related Work}
\subsection{Geometry-only Point Cloud Completion}

Point-based completion methods operate directly on unordered point sets, avoiding the quantization artifacts and computational overhead of voxel-based approaches \cite{GRnet}. A pioneering method, PCN \cite{PCN} introduced a coarse-to-fine framework, where a rough shape was first reconstructed and then refined. This paradigm has since been widely adopted and extended in various studies \cite{pcn-MVP, pcn-dualgenerator, pcn-p2p, pcn-temporal}, with subsequent works focusing on improving both the shape representation and refinement mechanisms.

Among these advancements, FoldingNet \cite{FoldingNet} introduced a folding-based decoder that maps 2D grids to 3D surfaces, capturing shape deformations effectively. TopNet \cite{Topnet} adopted a hierarchical rooted tree structure for multi-level shape generation, while MSN \cite{MSN} improved refinement through residual learning. Further improvements were introduced by PF-Net \cite{PFnet}, which enhanced completion by integrating multi-scale feature encoding with adversarial training, and GRNet \cite{GRnet}, which utilized a 3D grid-based representation to leverage CNNs while preserving structural information. 

Recently, transformer-based approaches have demonstrated promising advancements. PoinTr \cite{Pointr} and SnowflakeNet \cite{SnowflakeNet}, reformulated point cloud completion as a set-to-set translation problem, iteratively refining point distributions to recover fine details.

While these methods achieve promising results, they rely solely on geometric reasoning from partial inputs, limiting their ability to recover missing details effectively. To address this, cross-modal approaches have been explored, leveraging additional image information to improve point cloud completion.

\subsection{Cross-modal Point Cloud Completion}

Cross-modal point cloud completion aims to integrate visual cues from RGB images to compensate for missing geometric information in partial point clouds. Early approaches such as ViPC \cite{ViPC} and CSDN \cite{CSDN} explored shallow fusion strategies. ViPC generated a coarse shape from the image and refined it by directly concatenating image and point cloud features. CSDN enhanced this idea by proposing IPAdaIN to modulate point cloud features with image priors, followed by pixel-level alignment. While these methods provided initial improvements, their limited feature interaction often failed to capture rich cross-modal correspondences and struggled with structural inconsistencies in complex shapes.

To enable deeper multi-modal interaction, recent methods such as XMFNet \cite{XMFNet}, CDPNet \cite{CDPNet}, and EGIINet \cite{EGIINet} employed more advanced attention-based fusion strategies. XMFNet stacked self- and cross-attention layers to directly fuse features in latent space, but its extensive attention blocks introduced significant computational overhead and optimization difficulty. CDPNet adopted a two-stage refinement framework—predicting global shape from image features followed by patch-wise local refinement—which balances semantic completeness and geometric detail. EGIINet further unified 2D and 3D encoding with a guided interaction mechanism that emphasizes critical image regions, achieving strong cross-modal alignment and completion accuracy.

Building upon these advancements, our method introduces a hierarchical cross-modal fusion framework that combines the benefits of structural guidance and attention efficiency. Unlike ViPC and CSDN, which perform early fusion through simple concatenation or style modulation, we enable deep feature interaction through hierarchical attention. Compared to XMFNet and EGIINet, we apply attention selectively at different levels of geometric abstraction, improving both computational efficiency and reconstruction quality by preserving global structure and fine-grained details.

\begin{figure*}[!t]
\centering
\vspace{-0.24cm}
\centering
\includegraphics[width=0.9\textwidth]{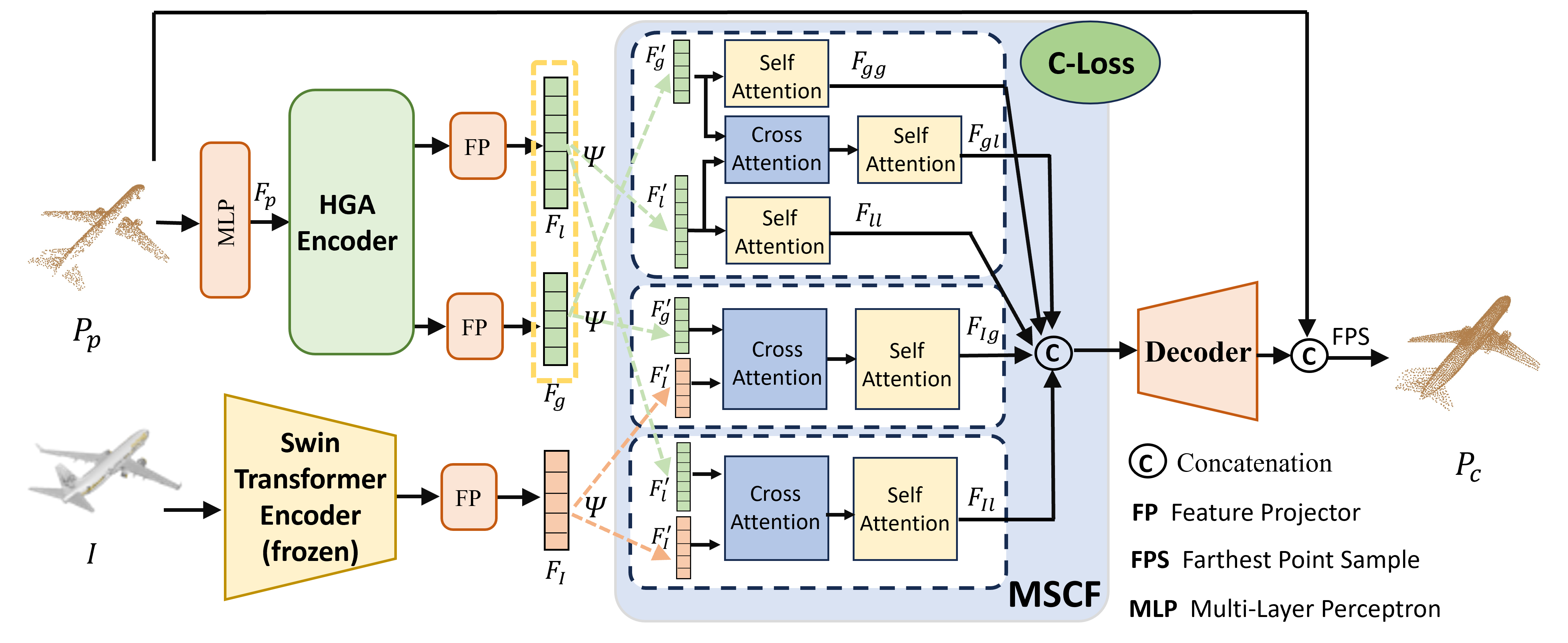}
\vspace{-0.1cm}
\caption{Architecture of HGACNet framework. Given a partial point cloud $P_p$ and an image $I$, the HGA encoder and Swin Transformer extract geometric and visual features, respectively. The MSCF module performs multi-scale fusion by aligning geometric features (${F_g}$ and ${F_l}$) and image features ${F_I}$ through cross-attention and self-attention operations, guided by the C-Loss. Finally, the fused features are decoded to generate the complete point cloud $P_c$.}
\label{fig2}
\vspace{-0.3cm}
\end{figure*}

\section{Method}

Given a partial point cloud $P_p \in \mathbb{R}^{N_p \times 3} $ and a corresponding image $I \in \mathbb{R}^{H \times W \times 3} $ ,our goal is to accurately predict the dense complete point cloud $P_c \in \mathbb{R}^{N_c \times 3} $ accurately, where $N_p$ and  $N_c$ denote the number of points in the partial and complete point clouds, respectively. Our framework first encodes $P_p$ using the proposed HGA encoder, which employs graph attention-based downsampling to extract both global feature $F_g \in \mathbb{R}^{N_g \times D_g} $ and local feature $F_l \in \mathbb{R}^{N_l \times D_l} $, where $N_g$ and $N_l$ are the number of selected key points at different hierarchy levels, and $D_g$ and $D_l$ are their corresponding feature dimensions. Meanwhile, the image $I$ is processed using a pre-trained and frozen Swin Transformer \cite{SwinTransformer} to obtain image features $F_I \in \mathbb{R}^{N_I \times D_I} $, where $N_I$ is the number of image features and $D_I$ is the feature dimension. Next, the MSCF module integrates the hierachical point cloud features $F_g$ and $F_l$ with image features $F_I$ by leveraging both cross-attention and intra-modal interactions, while a C-Loss is employed to enhance cross-modal consistency. The resulting fused features are then concatenated and passed through a decoder to reconstruct the completed point cloud. An overview of our architecture is illustrated in Fig.~\ref{fig2}.

\begin{figure}[!t]
\centering
\vspace{-0.0cm}
\centering
\includegraphics[width=0.45\textwidth]{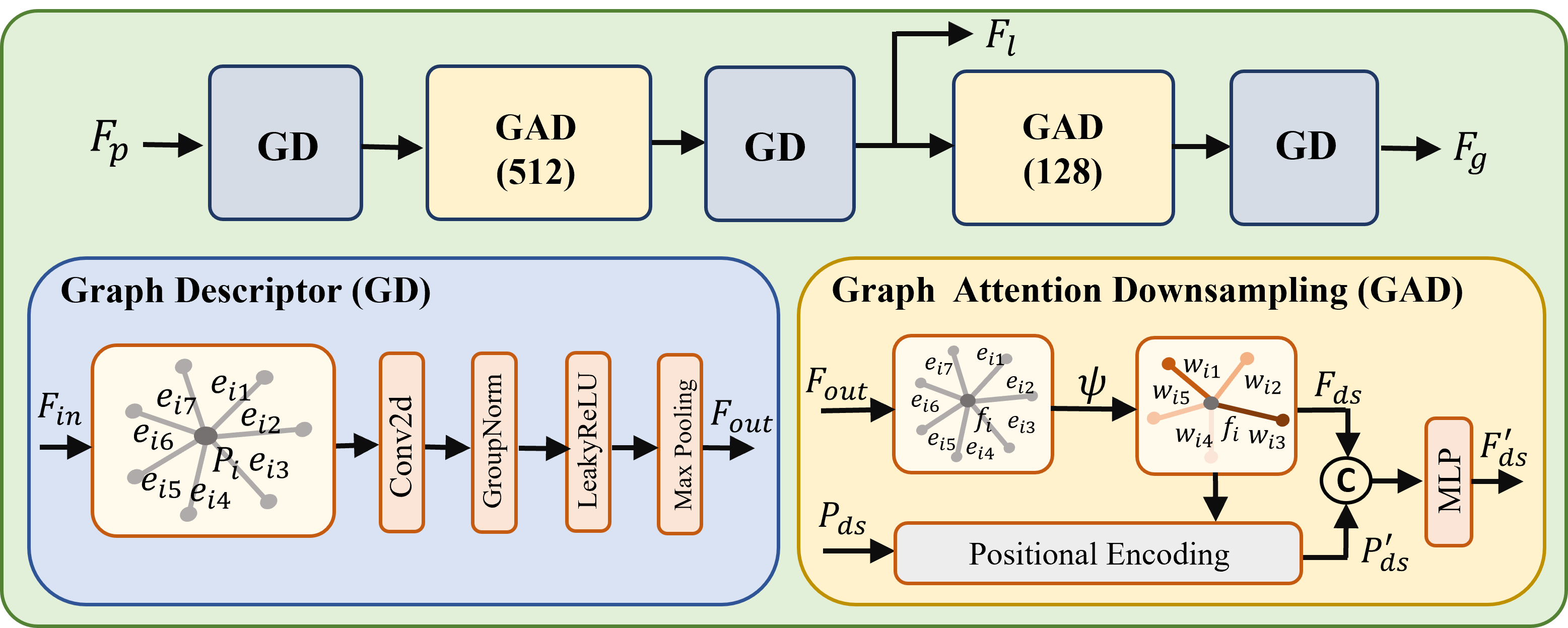}
\vspace{-0.1cm}
\caption{Structural design of the HGA encoder. It extracts hierarchical features via stacked Graph Descriptor and Graph Attention Downsampling modules, capturing both local details and global structure.}

\label{fig3}
\vspace{-0.3cm}
\end{figure}

\subsection{Hierarchical Encoder}
Both global structural information and local geometric details are critical for effective point cloud completion. In our approach, we design an HGA Encoder to extract hierachical point cloud features while employing a Swin Transformer to encode image features. The combination of these two modalities allows our model to leverage complementary information for enhanced reconstruction.

\subsubsection{Point Cloud Encoding with HGA}
 HGA Encoder is designed to adaptively select key points for feature extraction. As shown in Fig.~\ref{fig3}, the encoding pipeline consists of two key stages: Graph Descriptor (GD), and Graph Attention Downsampling (GAD).

\textbf{GD Module:} This module encodes local geometric structures by transforming raw point cloud features into a rich, structured representation. Given an input point cloud $P_p \in \mathbb{R}^{N_p \times 3}$, we first generate point-wise features $F_p \in \mathbb{R}^{N_p \times D}$ using a shared MLP. A local graph is then constructed for each point $p_i$ by connecting it to its $K_1$ nearest neighbors, allowing the network to capture local topological and spatial relationships. The edge features $e_{ij}$ between a point  $p_i$ and its neighbors $p_j$ are formulated as:
\setlength\abovedisplayskip{3pt}
\setlength\belowdisplayskip{3pt}
\begin{equation}
    e_{ij} = \phi ( p_i - p_j, f_i, f_j ),
\end{equation}
where $f_i$ and $f_j$ are the features of $p_i$ and $p_j$, respectively. $\phi(\cdot)$ is a learnable function that encodes both relative positional offsets and feature correlations within the local graph.

To extract discriminative representations from local structures, we employ a graph-based convolutional pipeline that progressively refines feature embeddings. The encoded edges are first processed by a 2D convolutional layer to expand the local receptive field and capture higher-order geometric cues. Group Normalization (GroupNorm) is applied to stabilize training by normalizing features across channel groups, followed by a LeakyReLU activation to introduce non-linearity. Finally, Max Pooling aggregates local neighborhood information, yielding compact feature embeddings that encode essential geometric context for downstream graph-based downsampling.

\textbf{GAD Modules:} A graph attention-based downsampling strategy is employed to adaptively select critical points while retaining structural details.The GAD module assigns importance scores to each node based on attention weights:
\setlength\abovedisplayskip{1pt}
\setlength\belowdisplayskip{1pt}
\begin{equation}
    w_i = \psi (f_i),
\end{equation}
where $\psi(\cdot)$ is a learnable scoring function. Points with the highest scores are selected to form a downsampled $P_p$ with $N_d$ points. This process is performed hierarchically, first reducing the point set to 512 nodes, then to 128 nodes, ensuring a local and global encoding of shape geometry.

To enhance spatial awareness, positional encoding is incorporated into the selected features:
\setlength\abovedisplayskip{1pt}
\setlength\belowdisplayskip{1pt}
\begin{equation}
    F'_{\text{ds}} = \text{MLP}\left( F_{ds} \oplus \gamma(P_{ds}) \right),
\end{equation}
where $\gamma(\cdot)$ is a sinusoidal encoding function, and $\oplus$ represents channel-wise concatenation.

Finally, the features obtained from different hierarchy levels are projected into a unified latent space through independent feature projection modules. The resulting local feature set $F_l$ preserves fine-grained geometric details, while the global feature set $F_g$ captures high-level semantic structures, laying the foundation for subsequent cross-modal fusion.

\subsubsection{Image Encoding with Swin Transformer}
To complement the point cloud representation, we employ a pretrained and frozen Swin Transformer \cite{SwinTransformer} as the image encoder, due to its ability to model long-range dependencies while preserving local spatial structure. The hierarchical design of the Swin Transformer naturally aligns with our point cloud encoder, capturing multi-scale visual features through window-based self-attention and shifted window mechanisms. Although not the focus of our contributions, this strong visual backbone facilitates effective alignment between 2D global context and 3D structural information during cross-modal fusion.

\subsection{Multi-scale Cross-Modal Fusion}
To effectively integrate point cloud and image features, we propose a hierarchical cross-modal fusion strategy that aligns local and global information across both modalities. Our method consists of two key components: (1) Cross-Modal cross-attention fusion, which enables fine-grained feature interaction between local and global features, and (2) contrastive learning, which enforces semantic consistency between the two modalities.

\subsubsection{Cross-Modal Cross-Attention Fusion}
As shown in the MSCF module of Fig.~\ref{fig2}, we design a hierarchical cross-modal fusion strategy that jointly exploits geometric and visual features at multiple scales.
Given global point cloud features $F_g$ and local point cloud features $F_l$ extracted by the HGA encoder, along with image features $F_I$ from the Swin Transformer, we first project all features into a unified latent space via linear transformations:
\setlength\abovedisplayskip{3pt}
\setlength\belowdisplayskip{3pt}
\begin{equation}
    F'_g = \Psi_g(F_g), \quad
    F'_l = \Psi_l(F_l), \quad
    F'_I = \Psi_I(F_I),
\end{equation}
where $\Psi(\cdot)$ denotes a learnable transformation function.

To enable rich context-aware interaction, we employ both self- and cross- attention across different hierarchical levels:
\begin{itemize}
    \item Self-attention is applied within $F'_g$ and $F'_l$ to model intra-modal dependencies, producing $F_{gg}$ and $F_{ll}$.
    \item Cross-attention is conducted between $F'_g$ and $F'_l$, allowing global and local features to mutually refine each other, resulting in $F_{gl}$.
    \item Cross-modal attention aligns point cloud features ($F'_g$, $F'_l$) with visual features $F'_I$, generating $F_{Ig}$ and $F_{Il}$ respectively.
\end{itemize}

Each attention operation is defined as:
\setlength\abovedisplayskip{1pt}
\setlength\belowdisplayskip{1pt}
\begin{equation}
    A = \text{Softmax} \left(\frac{QK^\top}{\sqrt{D}}\right), \quad \text{Output} = A V,
\end{equation}
where $Q$, $K$, and $V$ are projected queries, keys, and values from the respective feature sets.

\begin{figure}[!t]
\centering
\vspace{-0.0cm}
\centering
\includegraphics[width=0.45\textwidth]{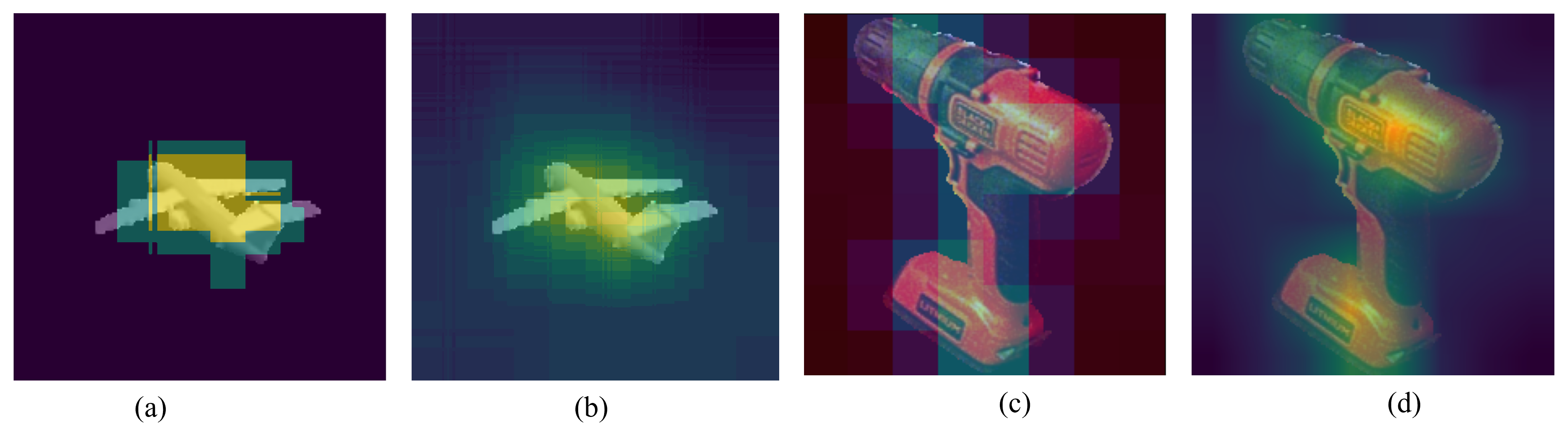}
\vspace{-0.1cm}
\caption{Visualization of cross-modal attention between point cloud and image features. (a) and (c) show global-to-image attention maps, while (b) and (d) show local-to-image attention. Global features capture the object structure, while local features focus on fine object details, demonstrating effective hierarchical alignment.}
\label{fig_attention}
\vspace{-0.3cm}
\end{figure}

To better understand how cross-modal alignment is achieved, we visualize the attention weights between point cloud and image features at different levels. As shown in Fig.~\ref{fig_attention}, global features primarily focus on the object’s overall shape and structure, helping to preserve completeness, while local features concentrate on semantically important details such as edges or functional parts. This behavior confirms the effectiveness of our hierarchical attention mechanism in aligning geometric and visual information at multiple scales.

Finally, the fused feature sets $\{F_{gg}, F_{gl}, F_{ll}, F_{Ig}, F_{Il}\}$ are concatenated and passed to the decoder for point cloud reconstruction.  
This hierarchical and cross-modal fusion design enables the network to simultaneously preserve fine-grained geometric details, maintain structural continuity, and leverage complementary semantic information from the image modality.

\subsubsection{Contrastive Learning for Cross-Modal Alignment}
To further strengthen feature consistency between point clouds and images, we proposed a C-Loss that aligns their global feature representations. Given a batch of global features from the point cloud $F_g$ and the corresponding image $F_I$, we encourage each point cloud feature to be maximally similar to its paired image feature while remaining dissimilar to other samples in the batch. This objective is achieved using an InfoNCE-based formulation:
\setlength\abovedisplayskip{3pt}
\setlength\belowdisplayskip{3pt}
\begin{equation}
\resizebox{0.42\textwidth}{!}{$
\begin{aligned}
\mathcal{L}_{\text{contrast}} = - \frac{1}{2} \Bigg[ 
    \sum_{i=1}^{B} \log \frac{\exp \left( \text{sim}(F_g^{(i)}, F_I^{(i)}) / \tau \right)}
    {\sum_{j=1}^{B} \exp \left( \text{sim}(F_g^{(i)}, F_I^{(j)}) / \tau \right)}  \\
    +  
    \sum_{i=1}^{B} \log \frac{\exp \left( \text{sim}(F_I^{(i)}, F_g^{(i)}) / \tau \right)}
    {\sum_{j=1}^{B} \exp \left( \text{sim}(F_I^{(i)}, F_g^{(j)}) / \tau \right)} 
\Bigg],
\end{aligned}
$}
\end{equation}
where $\text{sim}(\cdot, \cdot)$ denotes cosine similarity between feature vectors, and $\tau$ is a temperature scaling factor.

By jointly optimizing the C-Loss alongside the completion objective, our method promotes semantically aligned representations across modalities, thereby improving the coherence and accuracy of the completed point clouds.

 \subsection{Decoder}
 To reconstruct the complete point cloud from the fused features, we adopt a decoder architecture similar to that of XMFNet \cite{XMFNet}, which effectively transforms the integrated representation into final 3D coordinates. The fused features, incorporating both geometric and visual information, are directly passed through the decoder to predict the complete point cloud.Specifically, the decoder estimates a dense set of points that approximates the missing regions, which is then concatenated with a subsampled version of the input partial point cloud using farthest point sampling (FPS). This strategy allows the network to preserve both the observed geometry and the inferred structures, ensuring global completeness and local detail.

\subsection{Loss Function}
During the training phase, we employ two key loss functions to ensure the accuracy and structural consistency of the generated point clouds: the Chamfer Distance with $L_2$ norm (L2-CD) and the C-Loss. These losses work together to minimize geometric discrepancies while aligning cross-modal features for better reconstruction.
The CD with $L_2$ norm is defined as:
\setlength\abovedisplayskip{3pt}
\setlength\belowdisplayskip{3pt}
\begin{equation}
\textstyle
\resizebox{0.42\textwidth}{!}{$
\begin{aligned}
    \mathcal{L}_{\text{CD}_2}({\bf{P}},{\bf{Q}}) = \frac{1}{{\left| {\bf{P}} \right|}}\sum\limits_{x \in {\bf{P}}} {\mathop {\min }\limits_{y \in {\bf{Q}}} } \left\| {x - y} \right\|_2^2   
    + 
    \frac{1}{{\left| {\bf{Q}} \right|}}\sum\limits_{x \in {\bf{Q}}} {\mathop {\min }\limits_{y \in {\bf{P}}} } \left\| {y - x} \right\|_2^2,
    \label{eq:L2Chamfer}
\end{aligned}
$}    
\end{equation}
where $\| \cdot \|_2^2$ represents the squared $L_2$ norm, ensuring bidirectional alignment between the predicted and ground truth point clouds by penalizing point-wise discrepancies.

The final loss function is formulated as:
\setlength\abovedisplayskip{3pt}
\setlength\belowdisplayskip{3pt}
\begin{equation}
\textstyle
\resizebox{0.35\textwidth}{!}{$
    \mathcal{L}_{\text{total}} = \lambda_{\text{CD}_2} \mathcal{L}_{\text{CD}_2} + \lambda_{\text{contrast}} \mathcal{L}_{\text{contrast}},
$}   
\end{equation}
where $\lambda_{\text{CD}}$ and $\lambda_{\text{contrast}}$ balance the influence of CD and C-Loss. This combined objective enables our model to generate structurally accurate and semantically aligned point clouds. In our experiments, we set $\lambda_{\text{CD}} = 0.8$ and $\lambda_{\text{contrast}} = 0.2$, which provided a good trade-off between reconstruction accuracy and cross-modal consistency.

\begin{table*}[!t]
  \centering
  \captionsetup{font=normalsize,skip=2pt}
  \caption{Mean CD per point on ShapeNet-ViPC dataset (CD $\times 10^{-3} \downarrow$).}
  \label{tab:CDresult}
  \fontsize{10.5}{10.5}\selectfont 
  \begin{tabular}{cccccccccc}
    \toprule
    \textbf{Methods} & \textbf{Avg} & Airplane & Cabinet & Car & Chair & Lamp & Sofa & Table & Watercraft \\
    \midrule
    PCN         & 5.619 & 4.246 & 6.409 & 4.840 & 7.441 & 6.331 & 5.668 & 6.508 & 3.510 \\
    GRNet       & 3.171 & 1.916 & 4.468 & 3.915 & 3.402 & 3.034 & 3.872 & 3.071 & 2.160 \\
    TopNet      & 4.976 & 3.710 & 5.629 & 4.530 & 6.391 & 5.547 & 5.281 & 5.381 & 3.350 \\
    FoldingNet  & 6.271 & 5.242 & 6.958 & 5.307 & 8.823 & 6.504 & 6.368 & 7.080 & 3.882 \\
    PoinTr      & 2.851 & 1.686 & 4.001 & 3.203 & 3.111 & 2.928 & 3.507 & 2.845 & 1.737 \\
    PointAttN   & 2.853 & 1.613 & 3.969 & 3.257 & 3.157 & 3.058 & 3.406 & 2.787 & 1.872 \\
    \midrule
    ViPC        & 3.308 & 1.760 & 4.558 & 3.183 & 2.476 & 2.867 & 4.481 & 4.990 & 2.197 \\
    CSDN        & 2.570 & 1.251 & 3.670 & 2.977 & 2.835 & 2.554 & 3.240 & 2.575 & 1.742 \\
    XMFnet      & 1.443 & 0.572 & 1.980 & 1.754 & 1.403 & 1.810 & 1.702 & 1.386 & 0.945 \\
    EGIINet     & 1.211 & 0.534 & 1.921 & 1.655 & 1.204 & \textbf{0.776} & 1.552 & 1.227 & 0.802 \\
    \midrule
    \textbf{HGACNet (Ours)} & \textbf{1.002} & \textbf{0.377} & \textbf{1.458} & \textbf{1.340} & \textbf{1.028} & 0.981 & \textbf{1.105} & \textbf{1.074} & \textbf{0.597} \\
    \bottomrule
  \end{tabular}
\end{table*}

\begin{table*}[!t]
  \centering
  \captionsetup{font=normalsize,skip=2pt}
  \caption{Mean F-Score on ShapeNet-ViPC dataset  @ 0.001~\textbf{$\uparrow$}.}
  \label{tab:F1Score}
  \fontsize{10.5}{10.5}\selectfont
  \begin{tabular}{cccccccccc}
    \toprule
    \textbf{Methods} & \textbf{Avg} & Airplane & Cabinet & Car & Chair & Lamp & Sofa & Table & Watercraft \\
    \midrule
    PCN         & 0.407 & 0.578 & 0.270 & 0.331 & 0.323 & 0.456 & 0.293 & 0.431 & 0.577 \\
    GRNet       & 0.601 & 0.767 & 0.426 & 0.446 & 0.575 & 0.694 & 0.450 & 0.639 & 0.704 \\
    TopNet      & 0.467 & 0.593 & 0.358 & 0.405 & 0.388 & 0.491 & 0.361 & 0.528 & 0.615 \\
    FoldingNet  & 0.331 & 0.432 & 0.237 & 0.300 & 0.204 & 0.360 & 0.249 & 0.351 & 0.518 \\
    PoinTr      & 0.683 & 0.842 & 0.516 & 0.545 & 0.662 & 0.742 & 0.547 & 0.723 & 0.780 \\
    PointAttN   & 0.662 & 0.841 & 0.483 & 0.515 & 0.638 & 0.729 & 0.512 & 0.699 & 0.774 \\
    \midrule
    ViPC        & 0.591 & 0.803 & 0.451 & 0.512 & 0.529 & 0.706 & 0.434 & 0.594 & 0.730 \\
    CSDN        & 0.695 & 0.862 & 0.548 & 0.560 & 0.669 & 0.761 & 0.557 & 0.729 & 0.782 \\
    XMFnet      & 0.796 & 0.961 & 0.662 & 0.691 & 0.809 & 0.792 & 0.723 & 0.830 & 0.901 \\
    EGIINet     & 0.836 & 0.969 & 0.693 & 0.723 & 0.847 & \textbf{0.919} & 0.756 & 0.857 & 0.927 \\
    \midrule
    \textbf{HGACNet (Ours)} & \textbf{0.887} & \textbf{0.983} & \textbf{0.789} & \textbf{0.805} & \textbf{0.895} & 0.896 & \textbf{0.879} & \textbf{0.894} & \textbf{0.954} \\
    \bottomrule
  \end{tabular}
\end{table*}

\section{Experiments}

\subsection{Datasets and Metrics}

In our experiments, we utilize the ShapeNet-ViPC dataset \cite{ViPC}, derived from ShapeNet and structured for point cloud completion tasks. This comprehensive dataset comprises 38,328 objects across 13 categories, including airplane, bench, cabinet, car, chair, monitor, lamp, speaker, firearm, sofa, table, cellphone, and watercraft. Each object is associated with 24 rendered images from different viewpoints, 24 corresponding partial point clouds generated with occlusion, and a complete ground-truth point cloud uniformly sampled with 2048 points from the mesh surface.

Following the ViPC dataset configuration, we select eight categories: airplane, cabinet, car, chair, lamp, couch, table, and watercraft. The training data comprises 31,650 objects (80\% of the selected categories), with the remaining 20\% reserved for testing. Each partial and complete point cloud contains 2048 points. For the associated image data, a resolution of 224 × 224 pixels is maintained.

In our evaluation, we employ the L2-CD, as defined in Eq. \ref{eq:L2Chamfer}, and F-Score metrics to assess the performance of our model, following the standard practices adopted in previous works \cite{ViPC, EGIINet}. 

The F-Score, on the other hand, serves as an evaluation metric that integrates both precision and recall, providing a comprehensive assessment of the matching degree between the predicted point cloud and the reference point cloud. It is defined as:
\setlength\abovedisplayskip{3pt}
\setlength\belowdisplayskip{3pt}
\begin{equation}
F(\mathbf{P}, \mathbf{Q}) = \frac{2  \mathbf{P}(d) \mathbf{Q}(d)}{\mathbf{P}(d) + \mathbf{Q}(d)},
\end{equation}
where $\mathbf{P}(d)$ and $\mathbf{Q}(d)$ represent the proportions of points in $\mathbf{P}$ and $\mathbf{Q}$ that are within a threshold distance $d$ from the opposite set. The F-Score accounts for both the accuracy of surface reconstruction and the ability to capture the fine details, emphasizing how well the generated point cloud approximates the reference structure. The calculation of distances  adheres to the same principles applied in computing the L2-CD.

\subsection{Implementation Details}
The proposed framework is implemented in PyTorch and trained on an Nvidia 4090 GPU. Class-specific training is performed for all models, using the Adam optimizer for roughly 400 epochs with a batch size of 32. The learning rate is initialized to 0.1 and reduced by a factor of 10 at epoch 50, 80, 120 and 200.

\subsection{Results on ShapeNet-ViPC Dataset}

We evaluate our method on the ShapeNet-ViPC dataset \cite{ViPC}, and compare it quantitatively against existing cross-modal point cloud completion approaches \cite{ViPC, CSDN, XMFNet, EGIINet}, all trained under the same data conditions. In addition, we reference results from recent point-based completion methods \cite{GRnet, Topnet, FoldingNet, Pointr, pointattn}, as reported in \cite{XMFNet} to provide a broader context for comparison. The L2-CD and F1-score metrics are summarized in Table~\ref{tab:CDresult} and Table~\ref{tab:F1Score}, respectively.

Moreover, despite using a similar decoding strategy as prior works, our model achieves substantial improvements, with an average L2-CD reduction of 17\% compared to EGIINet and 31\% compared to XMFNet. This demonstrates that the proposed hierarchical cross-modal design effectively enhances structural precision and semantic completeness from partial observations.

\begin{figure*}[!t]
\centering
\vspace{0.04cm}
\centering
\includegraphics[width=0.8\textwidth]{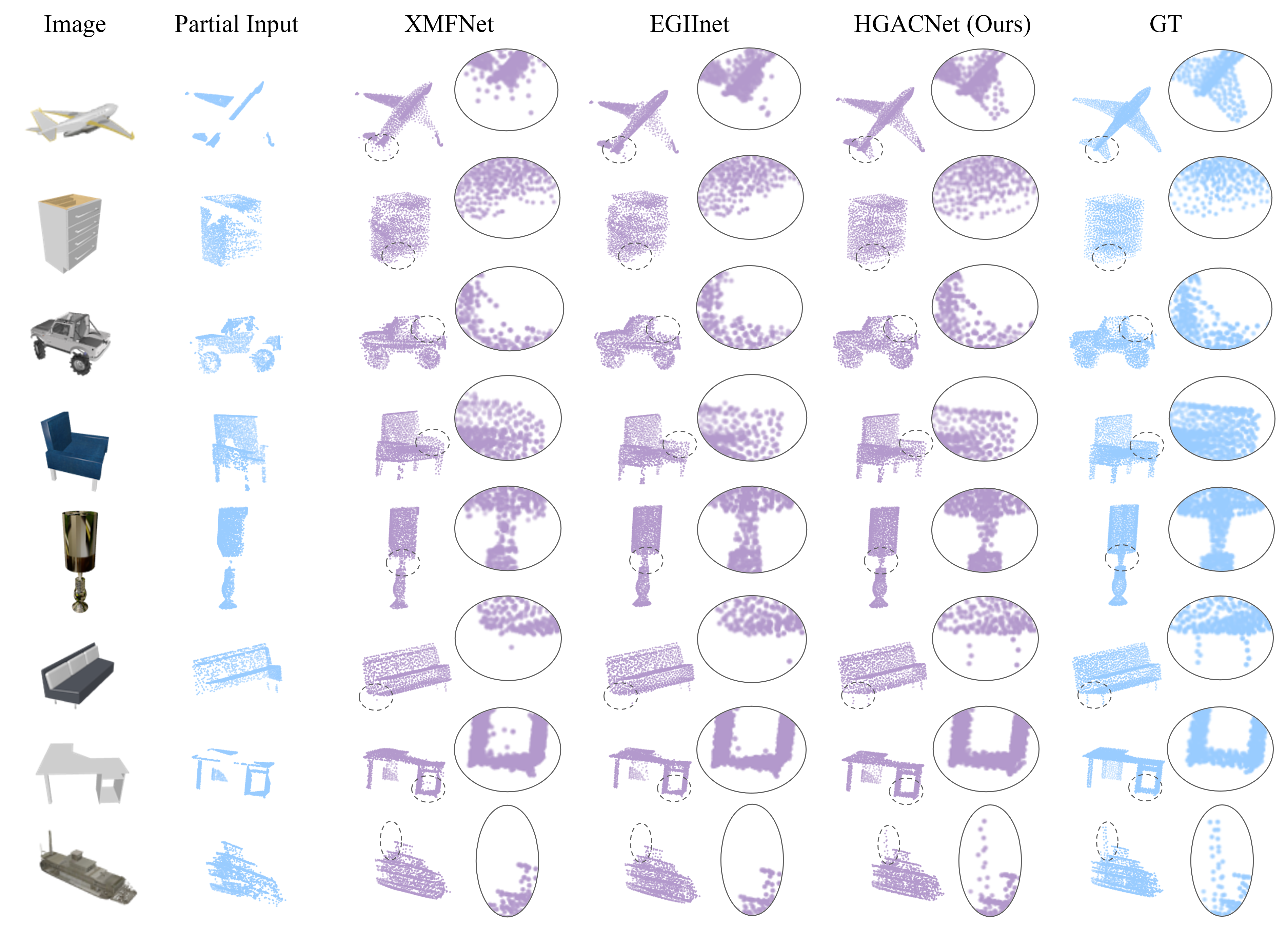}
\vspace{-0.4cm}
\caption{Qualitative comparison on the ShapeNet-ViPC dataset. Each row shows one sample, including the RGB image, partial input, and completion results from XMFNet, EGIINet, and our HGACNet, followed by the ground truth. HGACNet reconstructs more faithful and detailed shapes than the baselines, particularly excelling in thin and structurally complex regions. Key details are zoomed in for comparison.}
\label{fig4}
\vspace{-0.6cm}
\end{figure*}

Fig.~\ref{fig4} presents qualitative comparisons with XMFNet and EGIINet. As shown, HGACNet reconstructs finer structural details, such as airplane tail wings, cabinet drawers,the legs of sofas and tables, and the slender components of watercraft, where other methods often produce incomplete or distorted shapes. These improvements highlight the advantages of our hierarchical feature encoding and multi-scale cross-modal fusion in preserving fine-grained geometric detals.

\subsection{Ablation Studies}

To better understand the contribution of each module in our proposed framework, we conduct a series of ablation experiments focusing on four critical components: (1) the role of local feature extraction within the HGA encoder, (2) the effectiveness of the MSCF module in integrating geometric and visual information, (3) the impact of the proposed C-Loss for cross-modal alignment, and (4) the importance of incorporating image inputs as complementary priors. These ablations allow us to isolate and quantify how each design choice contributes to the overall performance of our model.

\textbf{Ablation on Local Feature Extraction.}
While most existing completion networks predominantly rely on global shape features, our HGA encoder is designed to extract both global context and local geometric details to enhance reconstruction fidelity. To assess the specific contribution of local features, we remove the local feature extraction module and retain only the global branch. As shown in Table~\ref{tab:ablation}, this modification leads to a noticeable performance drop, particularly in reconstructing fine structures. The results confirm that the incorporation of local signals significantly improves the ability of the model to capture detailed geometry and contributes to more complete and structurally coherent predictions.

\textbf{Ablation on MSCF.}
To examine its effectiveness, we replace the MSCF module with a simple feature concatenation strategy, where image and point cloud features are directly concatenated without interaction. As reported in Table~\ref{tab:ablation}, the performance drops significantly across all categories, with L2-CD increasing by 67\% on airplanes and 21\% on cars. This validates the necessity of explicit attention-based feature interaction for capturing meaningful cross-modal correspondences and improving structural completeness. We adopt one cross-attention and one self-attention layer at each scale to balance expressiveness and computational cost. Empirically, adding more layers only marginally improves performance but significantly increases inference time, which is critical for real-world deployment.

\textbf{Ablation on C-Loss.}
We proposed the C-Loss to enforce alignment between image and point cloud features in the latent space. To quantify its impact, we remove the C-Loss from the training objective. This leads to consistent performance degradation, especially on objects with complex shapes (e.g., Watercraft: 0.793 vs. 0.653). The results confirm that C-Loss effectively reduces the modality gap and encourages better feature correspondence, resulting in more accurate and coherent completions.

\textbf{Ablation on Image Input.}
To evaluate the importance of visual information, we conduct a variant that excludes the image input, making the network rely solely on geometric features from partial point clouds. As shown in Table~\ref{tab:ablation}, the absence of image guidance results in a noticeable performance drop, indicating the model's reduced capacity to infer unobserved geometry. This demonstrates that image priors significantly enhance the model’s ability to complete partial scans with higher fidelity, especially for categories with complex or thin structures.

\begin{table}[!t]
  \centering
  \captionsetup{font=normalsize,skip=2pt}
  \caption{Results of Ablation Studies (CD $\times 10^{-3}\downarrow$).}
  \label{tab:ablation}
  \fontsize{11}{11}\selectfont
  \resizebox{0.48\textwidth}{!}{ 
  \begin{tabular}{@{}ccccc@{}}
    \toprule
    Methods & Airplane & Cabinet & Car & Watercraft \\
    \midrule
    HGACNet & \textbf{0.377}  & \textbf{1.458} & \textbf{1.340}  & \textbf{0.653} \\
    w/o localfeat & 0.546 & 1.711 & 1.533 & 0.800 \\
    w/o MSCF & 0.632  & 1.764 & 1.619 & 0.834 \\
    w/o C-Loss & 0.512  & 1.634 & 1.439 & 0.793 \\
    w/o image & 0.471 & 1.751 & 1.472 & 0.744 \\
    \bottomrule
  \end{tabular}
  }
\end{table}

\begin{table}[!t]
\centering
\caption{Comparison on Real-World YCB-Complete Dataset.}
\label{tab:realworld}
\fontsize{11}{11}\selectfont 
\renewcommand{\arraystretch}{1.3}
\resizebox{0.48\textwidth}{!}{
\begin{tabular}{>{\centering\arraybackslash}m{2.5cm}|c|c|c|c}
\hline
\raisebox{-1.6ex}{\text{Methods}} 
& \multicolumn{2}{c|}{\text{Known}} 
& \multicolumn{2}{c}{\text{Unknown}} \\
\cline{2-5}
& CD ($\times 10^{-3}$) & F-Score & CD ($\times 10^{-3}$) & F-Score \\
\hline
XMFNet & 1.672 & 0.822 & 9.736 & 0.731 \\
EGIINet & 1.533 & 0.851 & 8.251 & 0.300  \\
\textbf{HGACNet (Ours)} & \textbf{0.073} & \textbf{0.995} & \textbf{7.996} & \textbf{0.405} \\
\hline
\end{tabular}
}
\vspace{-0.6cm}
\end{table}

\subsection{Real-World Evaluation on YCB-Complete}

\begin{figure}[!t]
\centering
\vspace{-0.14cm}
\centering
\includegraphics[width=0.45\textwidth]{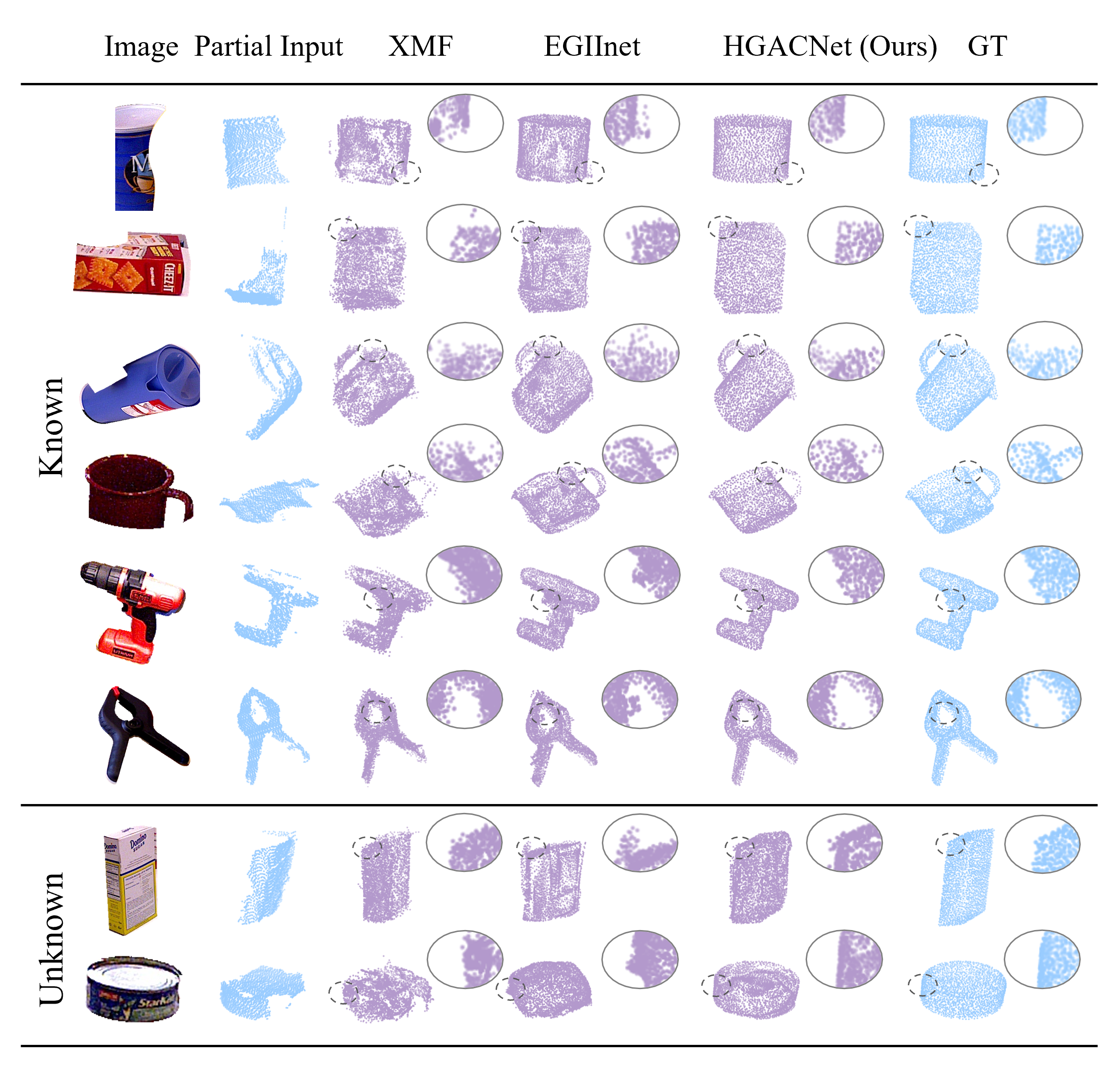}
\vspace{-0.3cm}
\caption{Qualitative comparison on the YCB-Complete dataset. HGACNet outperforms XMF and EGIINet by producing more complete and detailed reconstructions. Highlighted areas show its superiority in handling noise, occlusions, and unseen objects.}
\label{fig5}
\vspace{-0.2cm}
\end{figure}

To evaluate the real-world applicability of our method, we introduce YCB-Complete, a new object-level point cloud completion benchmark reorganized from the YCB-Video dataset \cite{ycbdataset}. The dataset contains 21 object categories, each with approximately 3,000 samples, including partial point clouds captured by a RealSense camera, corresponding RGB images, and aligned complete models.
Compared to synthetic datasets like ShapeNet-ViPC, YCB-Complete introduces real-world challenges such as sensor noise, occlusions, and pose misalignment, better reflecting practical robotic perception scenarios.

\textbf{Experimental Setup.} We divide the objects into 17 known and 3 unknown categories (\textit{004\_sugar\_box}, \textit{007\_tuna\_fish\_can}, and \textit{052\_extra\_large\_clamp}) to evaluate category-level generalization. Unknown objects share structural similarities with known classes, providing a realistic test of model adaptability. Following the ShapeNet-ViPC protocol, we adopt a split of  80\%/20\% train/test. Given the limited intra-category variation, training with only 20 epochs achieves stable convergence and competitive performance.

\textbf{Results and Analysis.} Table~\ref{tab:realworld} summarizes the quantitative results on the YCB-Complete dataset. HGACNet consistently outperforms XMFNet and EGIINet on both known and unknown categories, achieving the best results in both CD and F-score. The performance gap is more pronounced on known categories, where HGACNet produces highly accurate completions. On unknown categories, although the CD remains higher, this is largely due to the limited number of training classes in YCB-Complete, which restricts the diversity seen during training. Nevertheless, HGACNet still shows strong generalization, outperforming prior methods without any fine-tuning.

Fig.~\ref{fig5} presents qualitative comparisons among XMF, EGIINet, and our HGACNet. HGACNet achieves more accurate and complete reconstructions, especially in fine structures and object boundaries. Notably, it recovers the handle of the pitcher base, mug handles, and the thin edge of the power drill more accurately, where EGIINet and XMF often yield incomplete or distorted outputs. On unseen categories like \textit{004\_sugar\_box} and \textit{007\_tuna\_fish\_can}, HGACNet better preserves the global shape despite large missing regions. Highlighted areas show our method's advantages in structural integrity and generalization.

Overall, HGACNet shows strong adaptability to real-world sensory data, making it well-suited for downstream robotic tasks such as grasping, pose estimation, and scene interaction.

\section{Conclusion}
We present HGACNet, a hierarchical cross-modal framework for point cloud completion that integrates geometric structures from point clouds with visual priors from images. The proposed HGA encoder uses graph attention-based downsampling to adaptively select representative key points, capturing both global context and local geometric details. To enable effective cross-modal alignment, the MSCF module applies cross-attention to fuse image and point cloud features, while the proposed C-Loss further enhances consistency by narrowing the feature gap between modalities.

Extensive experiments on the ShapeNet-ViPC benchmark show that HGACNet achieves state-of-the-art performance, validating the effectiveness of its hierarchical encoding and multi-scale fusion strategy. Ablation studies confirm the contribution of each core component, including feature extraction, cross-modal fusion, and contrastive learning.

To evaluate real-world applicability, we constructed YCB-Complete, a benchmark derived from the YCB-Video dataset, featuring sensor noise and occlusions. HGACNet maintains strong performance under these conditions, highlighting its potential in robotic tasks such as object grasping, pose estimation, and scene interaction.

In future work, we plan to improve the adaptability of cross-modal fusion to better handle diverse scenarios, and explore unsupervised training strategies to reduce dependence on synthetic or fully labeled data.




\bibliographystyle{IEEEtran}
\bibliography{reference}

\begin{thebibliography}{10}
\providecommand{\url}[1]{#1}
\csname url@samestyle\endcsname
\providecommand{\newblock}{\relax}
\providecommand{\bibinfo}[2]{#2}
\providecommand{\BIBentrySTDinterwordspacing}{\spaceskip=0pt\relax}
\providecommand{\BIBentryALTinterwordstretchfactor}{4}
\providecommand{\BIBentryALTinterwordspacing}{\spaceskip=\fontdimen2\font plus
\BIBentryALTinterwordstretchfactor\fontdimen3\font minus \fontdimen4\font\relax}
\providecommand{\BIBforeignlanguage}[2]{{%
\expandafter\ifx\csname l@#1\endcsname\relax
\typeout{** WARNING: IEEEtran.bst: No hyphenation pattern has been}%
\typeout{** loaded for the language `#1'. Using the pattern for}%
\typeout{** the default language instead.}%
\else
\language=\csname l@#1\endcsname
\fi
#2}}
\providecommand{\BIBdecl}{\relax}
\BIBdecl

\bibitem{completeSUVey1}
K.~W. Tesema, L.~Hill, M.~W. Jones, M.~I. Ahmad, and G.~K. Tam, ``Point cloud completion: A survey,'' \emph{IEEE Transactions on Visualization and Computer Graphics}, vol.~30, no.~10, pp. 6880--6899, 2023.

\bibitem{completeSUVey2}
Z.~Zhuang, Z.~Zhi, T.~Han, Y.~Chen, J.~Chen, C.~Wang, M.~Cheng, X.~Zhang, N.~Qin, and L.~Ma, ``A survey of point cloud completion,'' \emph{IEEE Journal of Selected Topics in Applied Earth Observations and Remote Sensing}, vol.~17, pp. 5691--5711, 2024.

\bibitem{completeSUVey3}
B.~Fei, W.~Yang, W.-M. Chen, Z.~Li, Y.~Li, T.~Ma, X.~Hu, and L.~Ma, ``Comprehensive review of deep learning-based 3d point cloud completion processing and analysis,'' \emph{IEEE Transactions on Intelligent Transportation Systems}, vol.~23, no.~12, pp. 22\,862--22\,883, 2022.

\bibitem{applicationsimulating}
Z.~Liu, Z.~Chen, and W.-S. Zheng, ``Simulating complete points representations for single-view 6-dof grasp detection,'' \emph{IEEE Robotics and Automation Letters}, vol.~9, no.~3, pp. 2901--2908, 2024.

\bibitem{applicationpaint}
Y.~Zeng, D.~Zhang, S.-Y. Chien, H.~S. Tju, C.~Wiesse, F.~Cao, J.~Zhou, X.~Li, and I.-M. Chen, ``Task sensing and adaptive control for mobile manipulator in indoor painting application,'' \emph{IEEE/ASME Transactions on Mechatronics}, vol.~29, no.~4, pp. 2956--2963, 2024.

\bibitem{applicationkitting}
J.~Zhou, Y.~Zeng, H.~Dong, and I.-M. Chen, ``Discretizing so(2)-equivariant features for robotic kitting,'' in \emph{2024 IEEE/RSJ International Conference on Intelligent Robots and Systems (IROS)}, 2024, pp. 9502--9509.

\bibitem{Scpnet}
Z.~Xia, Y.~Liu, X.~Li, X.~Zhu, Y.~Ma, Y.~Li, Y.~Hou, and Y.~Qiao, ``Scpnet: Semantic scene completion on point cloud,'' in \emph{Proceedings of the IEEE/CVF conference on computer vision and pattern recognition}, 2023, pp. 17\,642--17\,651.

\bibitem{pc_upsampling2}
Y.~Rong, H.~Zhou, K.~Xia, C.~Mei, J.~Wang, and T.~Lu, ``Repkpu: Point cloud upsampling with kernel point representation and deformation,'' in \emph{Proceedings of the IEEE/CVF Conference on Computer Vision and Pattern Recognition (CVPR)}, June 2024, pp. 21\,050--21\,060.

\bibitem{GRnet}
H.~Xie, H.~Yao, S.~Zhou, J.~Mao, S.~Zhang, and W.~Sun, ``Grnet: Gridding residual network for dense point cloud completion,'' in \emph{European conference on computer vision}.\hskip 1em plus 0.5em minus 0.4em\relax Springer, 2020, pp. 365--381.

\bibitem{FoldingNet}
Y.~Yang, C.~Feng, Y.~Shen, and D.~Tian, ``Foldingnet: Point cloud auto-encoder via deep grid deformation,'' in \emph{2018 IEEE/CVF Conference on Computer Vision and Pattern Recognition}, 2018, pp. 206--215.

\bibitem{Pointr}
L.~An, P.~Zhou, M.~Zhou, Y.~Wang, and Q.~Zhang, ``Pointtr: Low-overlap point cloud registration with transformer,'' \emph{IEEE Sensors Journal}, vol.~24, no.~8, pp. 12\,795--12\,805, 2024.

\bibitem{CSDN}
Z.~Zhu, L.~Nan, H.~Xie, H.~Chen, J.~Wang, M.~Wei, and J.~Qin, ``Csdn: Cross-modal shape-transfer dual-refinement network for point cloud completion,'' \emph{IEEE Transactions on Visualization and Computer Graphics}, vol.~30, no.~7, pp. 3545--3563, 2024.

\bibitem{XMFNet}
E.~Aiello, D.~Valsesia, and E.~Magli, ``Cross-modal learning for image-guided point cloud shape completion,'' \emph{Advances in Neural Information Processing Systems}, vol.~35, pp. 37\,349--37\,362, 2022.

\bibitem{SwinTransformer}
Z.~Liu, Y.~Lin, Y.~Cao, H.~Hu, Y.~Wei, Z.~Zhang, S.~Lin, and B.~Guo, ``Swin transformer: Hierarchical vision transformer using shifted windows,'' in \emph{Proceedings of the IEEE/CVF international conference on computer vision}, 2021, pp. 10\,012--10\,022.

\bibitem{EGIINet}
H.~Xu, C.~Long, W.~Zhang, Y.~Liu, Z.~Cao, Z.~Dong, and B.~Yang, ``Explicitly guided information interaction network for cross-modal point cloud completion,'' in \emph{European Conference on Computer Vision}.\hskip 1em plus 0.5em minus 0.4em\relax Springer, 2024, pp. 414--432.

\bibitem{PCN}
W.~Yuan, T.~Khot, D.~Held, C.~Mertz, and M.~Hebert, ``Pcn: Point completion network,'' in \emph{2018 International Conference on 3D Vision (3DV)}, 2018, pp. 728--737.

\bibitem{pcn-MVP}
L.~Pan, T.~Wu, Z.~Cai, Z.~Liu, X.~Yu, Y.~Rao, J.~Lu, J.~Zhou, M.~Xu, X.~Luo \emph{et~al.}, ``Multi-view partial (mvp) point cloud challenge 2021 on completion and registration: Methods and results,'' \emph{arXiv preprint arXiv:2112.12053}, 2021.

\bibitem{pcn-dualgenerator}
P.~Shi, H.~Cheng, X.~Han, Y.~Zhou, and J.~Zhu, ``Dualgenerator: Information interaction-based generative network for point cloud completion,'' \emph{IEEE Robotics and Automation Letters}, vol.~8, no.~10, pp. 6627--6634, 2023.

\bibitem{pcn-p2p}
Z.~Zhang, Y.~Yu, and F.~Da, ``Partial-to-partial point generation network for point cloud completion,'' \emph{IEEE Robotics and Automation Letters}, vol.~7, no.~4, pp. 11\,990--11\,997, 2022.

\bibitem{pcn-temporal}
J.~Shi, L.~Xu, P.~Li, X.~Chen, and S.~Shen, ``Temporal point cloud completion with pose disturbance,'' \emph{IEEE Robotics and Automation Letters}, vol.~7, no.~2, pp. 4165--4172, 2022.

\bibitem{Topnet}
L.~P. Tchapmi, V.~Kosaraju, H.~Rezatofighi, I.~Reid, and S.~Savarese, ``Topnet: Structural point cloud decoder,'' in \emph{2019 IEEE/CVF Conference on Computer Vision and Pattern Recognition (CVPR)}, 2019, pp. 383--392.

\bibitem{MSN}
M.~Liu, L.~Sheng, S.~Yang, J.~Shao, and S.-M. Hu, ``Morphing and sampling network for dense point cloud completion,'' in \emph{Proceedings of the AAAI conference on artificial intelligence}, vol.~34, no.~07, 2020, pp. 11\,596--11\,603.

\bibitem{PFnet}
Z.~Huang, Y.~Yu, J.~Xu, F.~Ni, and X.~Le, ``Pf-net: Point fractal network for 3d point cloud completion,'' in \emph{2020 IEEE/CVF Conference on Computer Vision and Pattern Recognition (CVPR)}, 2020, pp. 7659--7667.

\bibitem{SnowflakeNet}
P.~Xiang, X.~Wen, Y.-S. Liu, Y.-P. Cao, P.~Wan, W.~Zheng, and Z.~Han, ``Snowflakenet: Point cloud completion by snowflake point deconvolution with skip-transformer,'' in \emph{Proceedings of the IEEE/CVF international conference on computer vision}, 2021, pp. 5499--5509.

\bibitem{ViPC}
X.~Zhang, Y.~Feng, S.~Li, C.~Zou, H.~Wan, X.~Zhao, Y.~Guo, and Y.~Gao, ``View-guided point cloud completion,'' in \emph{2021 IEEE/CVF Conference on Computer Vision and Pattern Recognition (CVPR)}, 2021, pp. 15\,885--15\,894.

\bibitem{CDPNet}
Z.~Du, J.~Dou, Z.~Liu, J.~Wei, G.~Wang, N.~Xie, and Y.~Yang, ``Cdpnet: cross-modal dual phases network for point cloud completion,'' in \emph{Proceedings of the AAAI Conference on Artificial Intelligence}, vol.~38, no.~2, 2024, pp. 1635--1643.

\bibitem{pointattn}
J.~Wang, Y.~Cui, D.~Guo, J.~Li, Q.~Liu, and C.~Shen, ``Pointattn: You only need attention for point cloud completion,'' in \emph{Proceedings of the AAAI Conference on artificial intelligence}, vol.~38, no.~6, 2024, pp. 5472--5480.

\bibitem{ycbdataset}
Y.~Xiang, T.~Schmidt, V.~Narayanan, and D.~Fox, ``Posecnn: A convolutional neural network for 6d object pose estimation in cluttered scenes,'' 2018.

\end{thebibliography}

\vfill

\end{document}